\documentclass{article}

\usepackage{arxiv}

\usepackage[utf8]{inputenc} 
\usepackage[T1]{fontenc}    
\usepackage{hyperref}       
\usepackage{url}            
\usepackage{booktabs}       
\usepackage{amsfonts}       
\usepackage{nicefrac}       
\usepackage{microtype}      
\usepackage{lipsum}		
\usepackage{graphicx}
\usepackage{natbib}
\usepackage{doi}

\title{Interactive Machine Learning: \\
A State of the Art Review}

\author{ 
  \href{https://orcid.org/0000-0000-0000-0000}{\includegraphics[scale=0.06]{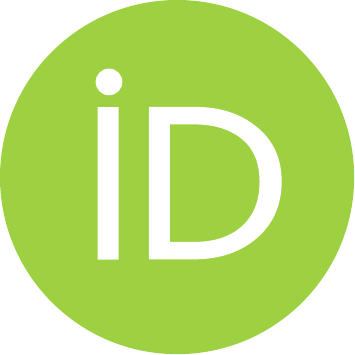}\hspace{1mm}Natnael A. WONDIMU}\thanks{ENIB, 945 Av. du Technopôle, 29280 Plouzané and Addis Ababa University, King George Street, Addis Ababa, Ethiopia.} \\
	Department of Artificial Intelligence\\
	Addis Ababa University\\
	Addis Ababa, Ethiopia \\
	\texttt{natnael.argaw@aau.edu.et} \\
	\And
	\href{https://orcid.org/0000-0000-0000-0000}{\includegraphics[scale=0.06]{orcid.pdf}\hspace{1mm}Ubbo VISSER \thanks{Department of Computer Science, University of Miami, 1320 S Dixie Hwy, Coral Gables, FL 33146, United States}} \\
	Department of Computer Science\\
	University of Miami\\
	Florida, USA \\
	\texttt{visser@cs.miami.edu} \\
	\And
	\href{https://orcid.org/0000-0000-0000-0000}{\includegraphics[scale=0.06]{orcid.pdf}\hspace{1mm}Cédric BUCHE \thanks{ENIB, 945 Av. du Technopôle, 29280 Plouzané - IRL CROSSING, Corner of North Terrace and Frome Rd, Adelaide SA 5000.} } \\
	IRL CROSSING, ENIB\\
	Brittany, France\\
	\texttt{cedric.buche@enib.fr} \\
}

\renewcommand{\shorttitle}{iML: A State of the Art Review}

\hypersetup{
pdftitle={Interactive Machine Learning: A State of the Art Review},
pdfsubject={q-bio.NC, q-bio.QM},
pdfauthor={Natnael A. Wondimu, Cédric BUCHE, Uboo VISSER},
pdfkeywords={interactive machine learning, exploratory machine learning, low resource learning, iML evaluation},
}

\begin{document}
\maketitle

\begin{abstract}
	Machine learning has proved useful in many software disciplines, including computer vision, speech and audio processing, natural language processing, robotics and some other fields. However, its applicability has been significantly hampered due its black-box nature and significant resource consumption. Performance is achieved at the expense of enormous computational resource and usually compromising the robustness and trustworthiness of the model. Recent researches have been identifying a lack of interactivity as the prime source of these machine learning problems. Consequently, interactive machine learning (iML) has acquired increased attention of researchers on account of its human-in-the-loop modality and relatively efficient resource utilization. Thereby, a state-of-the-art review of interactive machine learning plays a vital role in easing the effort toward building human-centred models. In this paper, we provide a comprehensive analysis of the state-of-the-art of iML. We analyze salient research works using merit-oriented and application/task oriented mixed taxonomy. We use a bottom-up clustering approach to generate a taxonomy of iML research works. Research works on adversarial black-box attacks and corresponding iML based defense system, exploratory machine learning, resource constrained learning, and iML performance evaluation are analyzed under their corresponding theme in our merit-oriented taxonomy. We have further classified these research works into technical and sectoral categories. Finally, research opportunities that we believe are inspiring for future work in iML are discussed thoroughly.
\end{abstract}


\section{Introduction}
Interactive machine learning (iML) is an active machine learning technique in which models are designed and implemented with human-in-the-loop manner. End-users participate in model building process by iteratively feeding training parameters, inspecting model outputs and providing feedback on intermediate results \cite{arendt2017chissl, amershi2014power, fails2003iml_2, jiang2019recent, Holzinger2018InteractiveML_9,boukhelifa2018evaluation_3,Shixia2017visual,francoise_2020}. 

The conception of iML dates back to the emergence of query learning where queries are used to learn unknown concepts \cite{ghai2021explainable, angluin1988queries, holzinger2016interactive, chung2020interactive,fogarty2008cueflik}. A model building component in iML framework interacts with Oracles by issuing queries for additional training data or feedback against its intermediate results. iML based methods mainly aspires to build robust \cite{huang2011adversarial, pappernot2017adv_ref4, carlini2017blackbox_ref9, szegedy2013blackbox_ref10, sun2018iterated, brendel2017blackbox_ref8, guo2019blackbox_ref11, cheng2019opt, guo2018low, porter2013interactive,
ma2019iml_46, das2020massif, slack2020fooling, emamjomeh2017general} models by trading-off accuracy for trust \cite{Stepano2019Explanatory_6, robert2017Robot_5, krause2016blackbox_ref12, RibeiroSG2016blackbox, Mozina2018explanatory_7, robert2017Robot_5, Holzinger2018InteractiveML_9, turchetta2019safe, berkenkamp2016bayesian, sui2018stagewise, van2011iml_ref14, liu2017iml_ref15, zhao2018iml_ref16, Muhlbacher2014blackbox_ref13} and low resource learning \cite{ambati2011lowresource_47, frazier2019improving, porter2013interactive, preuveneers2020resource_ref17,  holzinger2017glass, fails2003iml_2, amershi2011effective, tegen2020activity_16, amershi2012regroup_20, dzyuba2014interactive, li2018interactive, jain2020algorithmic, fiebrink2010wekinator_12,gillian2014gesture,schedel2011wekinating, diaz2019interactive, arendt2018interactive,arendt2017chissl}.

iML is becoming the center of machine learning research \cite{jiang2019recent, arendt2017chissl} as the need to operate in resource constrained environments and engaging the human-in-the-loop increased over time. Various iML systems have been designed and implemented to alleviate the long-standing setbacks of standard machine learning algorithms. Its impact has been studied in health \cite{Holzinger2018InteractiveML_9, wallace2012deploying, holzinger2016interactive, maadi2021review, holzinger2017glass, kose2015interactive, qian2011using, tyagi2018interactive}, agriculture \cite{de2020interactive, flutura2020interactive, liakos2018machine}, finance \cite{kose2015interactive, white1993survey}, politics  and sociology \cite{amershi2012regroup_20,gillies2019understanding_21}, military and robotics  \cite{robert2017Robot_5, berkenkamp2016bayesian, kyrarini2019robot}, cyber security \cite{chung2020interactive, huang2011adversarial, pappernot2017adv_ref4,carlini2017blackbox_ref9, brendel2017blackbox_ref8, guo2019blackbox_ref11, slack2020fooling, emamjomeh2017general, ilyas2018black, cheng2019improving}, education \cite{witten2002data_ref7,hollan1984steamer,zhou2020aispace2} and game and entertainment \cite{frazier2019improving, diaz2019interactive, fiebrink2010wekinator_12, schedel2011wekinating}. 

Moreover, iML has been analysed and built over the predominant standard machine learning algorithms such as SVM \cite{Stepano2019Explanatory_6, talbot2009ensemblematrix, tong2001support,tsai2009interactive, javidi2008new, tyagi2018interactive, talbot2009ensemblematrix}, Genetic Algorithms \cite{robert2017Robot_5, kim2000application,lai2011user, cho2002human, wang2005improved, poirson2020interactive}, Ant Colony \cite{holzinger2019iml_9, meng2021heterogeneous, yang2020interactive}  and some other algorithms \cite{tyagi2018interactive, frazier2019improving, fogarty2008cueflik, breve2020simple, ha2017neural}. Similarly, its capacity to uncover the details behind black-box models is presented in various research works \cite{robert2017Robot_5,Stepano2019Explanatory_6,van2011iml_ref14,liu2017iml_ref15,ma2019iml_46,Muhlbacher2014blackbox_ref13}. Most research focus ranges from extending standard machine learning models with interactive capabilities to formulating robust performance evaluation mechanisms.

Sate-of-the-art review gives insights on recent progresses on a specific subject matter and paves ways for more research. To this end, numerous surveys have been conducted on the application of iML to specific sectors. Architectural \cite{dudley2018review_4, chatzimparmpas2020survey,meza2019towards} and application/sector oriented \cite{jiang2019recent, liakos2018machine, maadi2021review} analysis has been significantly used techniques in the past. According to the meta-analysis \cite{meza2019towards, dudley2018review_4, maadi2021review, liakos2018machine, chatzimparmpas2020survey, jiang2019recent, amershi2014power, Shixia2017visual, teso2020challenges} we conducted, most surveys lack extensiveness and inclusiveness, leaving the various researches that constitute the state-of-the-art of iML untouched. In \cite{amershi2014power}, \cite{dudley2018review_4} and \cite{meza2019towards}, the advancements made towards designing usable interfaces for iML is discussed. A review of iML and interactive visualization has been conducted in \cite{Shixia2017visual} and \cite{meza2019towards}. Moreover, a review of iML inspired researches in medical and agricultural sectors is extensively studied in \cite{maadi2021review} and \cite{liakos2018machine} respectively. A recent task-oriented extensive survey on interactive machine learning is also presented in \cite{jiang2019recent}.

However, to the best of our knowledge, there has been a little extensive survey in the field of iML covering the state-of-the-art. Besides, no survey 
offer a merit-oriented and detailed state-of-the-art review. Therefore, the primary motivation of this paper is to summarize recent advancements in interactive machine learning by using merit-oriented taxonomy. We follow a similar approach like prior surveys but broadening our scope using merit-oriented taxonomy of iML.

In this paper, we review the state-of-the-art of interactive machine learning and thoroughly present research works in a merit-oriented taxonomy. While our main focus is on discussing iML contributions for robust, trustworthy and low resource machine learning, we have also presented issues related to iML performance evaluation. In section \ref{entry}, we discuss about the overall merit-oriented architecture of iML. In section \ref{robust}, adversarial attacks against black-box machine learning models are discussed. Moreover, it presents various iML inspired contributions aiming to combat adversarial attacks. We present researches conducted  to increase the explainability and interpretability as a way to achieve trustworthy machine learning in section \ref{trust}. Researches that employ human-in-the-loop for data processing and model building with the intention of lowering the data and computational resources are discussed in section \ref{lowresource}. In section \ref{imleval}, we discuss the subjective nature of iML and related efforts to address problems in evaluating it. Further analysis of findings and discussion is presented in \ref{Potential}. In the last section, a summary of the state-of-the-art review and potential research opportunities that are inspiring for future work on iML are discussed.

\section{Merit-Oriented Taxonomy of iML}
\label{entry}
\label{focus}
iML inspired contributions can be analysed from various perspectives. However, architectural \cite{dudley2018review_4, chatzimparmpas2020survey,meza2019towards} and application/sector oriented \cite{jiang2019recent, liakos2018machine, maadi2021review} analysis has been significantly used techniques in the past. However, according to the meta-analysis \cite{meza2019towards, dudley2018review_4, maadi2021review, liakos2018machine, chatzimparmpas2020survey, jiang2019recent, amershi2014power, Shixia2017visual, teso2020challenges} we conducted, most survey papers lack extensiveness and inclusiveness, leaving the various researches that constitute the state-of-the-art of iML untouched. Consequently, the employment of iML has been constrained to only predefined sectors or limited scope of the architecture. 

\begin{figure}[htbp]
  \centering
  \includegraphics[width=\linewidth]{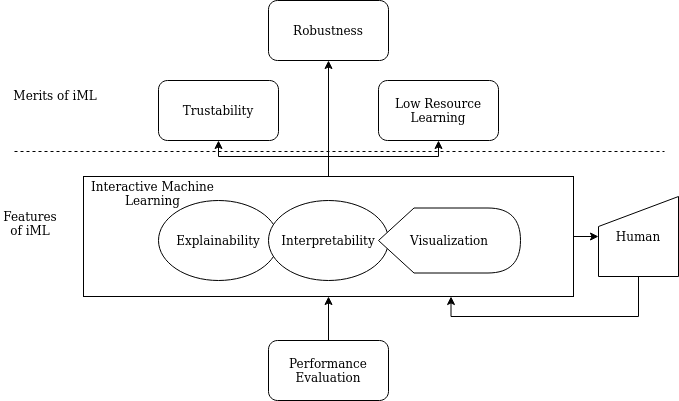}
  \caption{Merit-Oriented Architecture of Interactive Machine Learning}
  \label{imlcomp}
\end{figure}

To this end, we have thoroughly analysed the recent iML-inspired research works using merit-oriented taxonomy. After an extensive review of iML-inspired literature and untapped problems, we have categorized contributions into Robust Machine Learning \cite{huang2011adversarial, pappernot2017adv_ref4,carlini2017blackbox_ref9,szegedy2013blackbox_ref10, sun2018iterated, brendel2017blackbox_ref8, guo2019blackbox_ref11, cheng2019opt, guo2018low,
ma2019iml_46, das2020massif, slack2020fooling, emamjomeh2017general}, Trustworthy Machine Learning \cite{Stepano2019Explanatory_6, robert2017Robot_5, krause2016blackbox_ref12, RibeiroSG2016blackbox, Mozina2018explanatory_7, robert2017Robot_5, Holzinger2018InteractiveML_9, turchetta2019safe, berkenkamp2016bayesian, sui2018stagewise, van2011iml_ref14, liu2017iml_ref15, zhao2018iml_ref16, Muhlbacher2014blackbox_ref13}, and Low Resource Machine Learning
\cite{ambati2011lowresource_47, frazier2019improving, porter2013interactive, preuveneers2020resource_ref17, holzinger2017glass, fails2003iml_2, amershi2011effective, tegen2020activity_16, amershi2012regroup_20, dzyuba2014interactive, li2018interactive, jain2020algorithmic, fiebrink2010wekinator_12,gillian2014gesture,schedel2011wekinating, diaz2019interactive, arendt2018interactive, arendt2017chissl} based on their merit in the solution space. Besides, performance evaluation techniques that face bias because of the subjective nature of iML are analysed \cite{boukhelifa2018evaluation_3, krause2014infuse_ref19, ma2019iml_46, yang2019evaluating_45, krause2017workflow_ref18, felderer2021quality}. A high-level notional representation of merit-oriented architecture of iML is depicted in Figure \ref{imlcomp}. Our state-of-the-art review is channeled under these major merits of iML.

\section{Robust Machine Learning}
\label{robust}
Deep learning algorithms are known for their vulnerability to adversarial attacks \cite{szegedy2013intriguing}. Adversaries can craftily manipulate legitimate inputs, which may be imperceptible to human eye, but can force a trained model to produce incorrect outputs. This issue is directly related to the black-box and intricacy nature of deep learning \cite{pappernot2017adv_ref4, carlini2017blackbox_ref9,szegedy2013blackbox_ref10, sun2018iterated, tu2019autozoom, tramer2017ensemble, huang2011adversarial}.

We use \cite{papernot2016towards} and \cite{chakraborty2018adversarial} threat model to highlight the state-of-the-art of the adversarial attack. According to \cite{papernot2016towards}, adversarial threat model is comprised of the following three dimensions, namely: the attack surface, adversarial capabilities, and adversarial goals. The attack surface refers to the overall data processing pipeline of machine learning from input to output and then action. Evasion attack \cite{biggio2013evasion, kwon2018friend, kwon2018multi}, poisoning attack \cite{chen2017targeted, zhang2019poisoning, geiping2021doesn} and exploratory attacks \cite{pappernot2017adv_ref4, brendel2017blackbox_ref8, guo2019blackbox_ref11} are the main scenarios considered under the attack surface. The aforementioned attacks can further be dissected into training phase \cite{chen2017targeted, zhang2019poisoning, geiping2021doesn} and testing phase \cite{biggio2013evasion, kwon2018friend, kwon2018multi} attacks from the adversarial capabilities point of view. Data injection, data modification and logic corruption are among the strategies of training phase adversarial capability attacks. On the other hand, white-box \cite{moosavi2016deepfool,tramer2016stealing} and black-box \cite{papernot2016transferability, rosenberg2017generic} attacks are among the testing phase capability attacks. Adversarial goal attack, on the other hand, infer adversary from the incorrectness of the target model. It is also further classified as confidence reduction, misclassification, targeted misclassification and source/target misclassification.

Adversarial attacks have real impacts on the robustness of a deep learning and other standard machine learning methods \cite{wang2021adversarial}. Therefore, exploring potential adversarial attacks and building a robust machine learning has been the focus of machine learning researchers.

There are various techniques being studied to address the problem of adversarial attacks. Researchers in the iML domain derived a number of strategies both to showcase the impact of adversaries \cite{morris2020textattack} and tackle \cite{slack2020fooling, das2018adagio} adversarial attack strategies discussed in \cite{papernot2016towards}. The fundamental assumption is that assuring the explainability and interpretability of black-box models by having the human-in-the-loop reduce the vulnerability of machine learning models to adversarial attack \cite{rudin2019iml_ref5}.

iML has been used as a way to explain and explore model vulnerabilities to adversarial attacks as in \cite{ma2019iml_46} and \cite{das2020massif}. Specifically, \cite{ma2019iml_46} enables exploration and explanation of model vulnerabilities to test-phase or poisoning adversarial attacks from the perspective of models, data instances, features, and local structures. A white-box exploratory attack based approach, \cite{das2020massif}, interactively visualizes neurons and their connections inside a DNN that are strongly activated or suppressed by an adversarial attack. Massif provides both a high-level, interpretable overview of the effect of an attack on a DNN, and a low-level, detailed description of the affected neurons.

In addition to explaining potentially perturbated inputs and models, iML based methods such as \cite{slack2020fooling} propose frameworks that masks the discriminatory biases of black-box classifiers. This plays a vital role to compensate the effects of perturbated inputs on a given model.

In addition to letting the user detect potential adversarial attacks and managing its impacts, iML has also been used to directly engage users in the model building process. This helps to avoid both training phase and test phase adversarial attacks as the user will be there validating inputs and intermediate results. A graph based framework, \cite{emamjomeh2017general}, where user feedback is represented as edges and nodes are the models is a good example of such applications. It proposes Angluin’s equivalence query model \cite{angluin1988queries, angluin1992computational} and Littlestone’s online learning model \cite{littlestone1988learning} based on a robust machine learning framework that interactively learns models like classifiers \cite{angluin1988queries,angluin1992computational}, orderings/rankings of items \cite{joachims2002optimizing,radlinski2005query}, or clusterings of data points \cite{awasthi2014local,emamjomeh2018adaptive,balcan2008clustering}. In each iteration, the algorithm proposes a model, and the user either accepts it or reveals a specific mistake in the proposal. The feedback is correct only with probability $p > 1/2$ (and adversarially incorrect with probability 1 $ - p$), i.e., the algorithm must be able to learn in the presence of arbitrary noise.

Interactive machine learning has also been applied to tackle adversarial attacks on automatic speech recognition (ASR) models. For instance, \cite{morris2020textattack} designed a tool allowing experimentation with adversarial attacks and defenses on ASR. It implemented iML techniques and psychoacoustic principles to effectively eliminate targeted attacks. A similar and cloud-based tool to evaluate and compare state-of-the-art adversarial attacks and defenses for machine learning (ML) models is proposed in \cite{das2019mlsploit}.

\section{Trustworthy Machine Learning}
\label{trust}
To design and develop AI-based systems that users and the larger public can justifiably trust, one needs to understand how machine learning technologies impact trust \cite{toreini2020relationship}. The trustworthiness of AI-based systems is directly related to how the user is confident about the decisions made by the machine learning components. This may include its perception about both the intelligent model and knowledge \cite{yin2019understanding}.

Users understanding of a given domain may negatively affect the trustworthiness of a model built around that specific domain. As it clearly stated in \cite{honeycutt2020soliciting}, the act of providing feedback can affect user understanding of an intelligent system and its accuracy. User trust toward a given model and even the perception of model accuracy might lower; regardless of whether the system accuracy is improved in response to their feedback. However, trustworthiness of a model should be defined considering its relative ability, benevolence and integrity.

To this end, we use the Trust Antecedent (TA) framework \cite{mayer1995integrative} definition of trust. According to this framework a trustee is given trust if it is perceived to have the ability, benevolence, and integrity toachive the desired goal. However, a black-box machine learning model that doesn't show its implementation details can not comply with the principles of this framework, hence affecting its trustworthiness. Besides, the black-box nature of these models significantly affects their applicability in in decision sensitive areas like health, finance, autonomous vehicles, criminal justice, etc.

Consequently, enhancing the trustworthiness of machine learning algorithms comes with making the model building process interactive \cite{robert2017Robot_5, krause2016blackbox_ref12, maadi2021review}. Put succinctly, increasing the explainability and interpretability of machine learning by engaging user-in-the-loop increase its trustworthiness.

To this end, various explanatory frameworks that shows the implementation details of the model building process have been introduced. The essence of most exploratory algorithms is that intermediate results corresponding to a batch or even tuple are made subject to user-feedback whenever there is a variation between the predicted and the actual label.

In important contributions like \cite{Stepano2019Explanatory_6, van2011iml_ref14, Mozina2018explanatory_7}, the iML is set to interactively query labels whenever the intermediate model fails to predict the label for each data point. For every prediction, the model provides the predicted label and explanations for the prediction. However, the way the the model and feedback are could be visual or textual.

Visual explainer systems are designed considering a set of guidelines proposed in various research contributions. For instance, \cite{Muhlbacher2014blackbox_ref13} propose a set of guidelines for integrating visualization into machine learning algorithms through a formalized description and comparison. Specific to automated iterative algorithms, which are widely used in model optimization, these researchers recommend exposing APIs for visualization developers. Using high-resolution APIs, visualization developers access the internal iterations for a tighter integration of the user in the decision loop.

Accordingly, \cite{Stepano2019Explanatory_6} propose a framework called CAIPI is proposed. It use a model-agnostic explainer, LIME \cite{RibeiroSG2016blackbox}, as fundamental component to compute the explainer and present them to the user as interpretable (visual) artifacts. Similarly, \cite{van2011iml_ref14} propose a natural visual representation of decision tree structures where decision criterion is visualized in the tree nodes. BOOSTVis \cite{liu2017iml_ref15} and iForest \cite{zhao2018iml_ref16} also focus on explaining tree ensemble models through the use of multiple coordinated views to help explain and explore decision paths.

Models and feedback from users can also be represented using textual-explanatory systems. An argument based explanatory iML algorithm \cite{Mozina2018explanatory_7} propose a framework that exploits the usability of arguments to precisely axiomatize feedbacks both from the user and system side. It practically shows how to narrow the gap between domain experts and the machine learning model by engaging domain experts. Users provide feedbacks using a pair of reasons and outputs called arguments. Since arguments are generally presumptive and can't be untaken for a general set of predictions, the authors introduced prediction-level argument based explanations of decisions made by the learning model. Put succinctly, the training module generates initial features to perform prediction or classification. Whenever the learner notice major deviation between the desired and actual prediction, it consults the domain expert to provide feedback on the output (both the prediction and explanation). At this stage, the domain expert would be able to see the problem either in the predicted label or the rule yielding the outcome. Then, the user provides a set of arguments for each critical example (predictions with problem) and let the model to retrain keeping the feedback given from the user. Moreover, the paper dictated how the user should communicate with the machine learning component through a sequence of steps namely: Selection, Presentation, Argument formulation, Counter example detection, refinement, argument pruning. Another goal goal-oriented safe exploration algorithm that provably avoids unsafe decisions in real world problems is proposed in \cite{turchetta2019safe}.This framework takes suggested decisions as input and exploits regularity assumptions in terms of a Gaussian process prior in order to efficiently learn about their safety.

Exploratory machine learning has been applied to various decision sensitive sectors such as autonomous vehicles, health, and military. In \cite{robert2017Robot_5} used exploratory components to build trust among autonomous robots. Specifically, they analyzed the importance of using iML with neuro-evolutionary algorithms and conducted two separate tests were conducted on two autonomous environments; one with iML empowered ML algorithm and the other with black-box ML algorithms. The major problem they selected for this experiment is Search. Participants in this research engaged in building iML based search plan models. Finally, they were allowed to see and distinguish between the plans as iML generated or black-box generated. To this end, participants were able to choose iML based search plans for its effectiveness but with less trust relative to the black-box algorithm. The authors justified the preliminary knowledge of participants as the reason for their low degree of trust on the iML algorithm. In conclusion, the researchers convey the importance of using iML based neuro-evolutionary algorithms for searching problems of autonomous robots.

The application of iML in the health sector has also alleviated major computational trust related problems \cite{Holzinger2018InteractiveML_9, berg2019ilastik_8}. A biological image data analysis tool called ilastik \cite{berg2019ilastik_8} propose an iML framework that address the speed and usability requirements of machine learning by reasonably compromising model accuracy. Ilastik is able to formulate a feature space without the need to use bulk data as the other classical machine learning algorithms.

\section{Low Resource Machine Learning}
\label{lowresource}
Low resource machine learning is a process of building an analytical model employing optimal resource utilization techniques. However, most machine learning algorithms have tradeoff between accuracy and resource utilization. As it is stated in \cite{preuveneers2020resource_ref17}, the most accurate model might be prohibitively expensive to computationally evaluate on a resource constrained environment. Consequently, the problem of building accurate and high performance machine learning models has been achieved at the expense of resource (data and computing) utilization. Nonetheless, sufficient amount of data and computing resources are not always at the stake. Besides, some problems may also require to be run on low resource setting. For instance, the use of pervasive devices and robots to build model is one valid scenario.

To this end, interactive machine learning help scientists and engineers exploit more specialized data within their deployed environment in less time, with greater accuracy and fewer costs \cite{porter2013interactive}. Various iML inspired researches has been conducted in the past couple of years. In this section, we will be discussing iML inspired solutions that aim at enabling low resource machine learning and improving the speed of models.

We have categorized our analysis of iML for low resource learning into small data machine learning and pervasive machine learning. In the small data machine learning section, we discuss contributions that use small data to build models in a low resource environment. On the other hand, the application of iML in pervasive low resource environment is discussed in the pervasive machine learning section.

\subsection{Small Data Machine Learning}
Machine learning algorithms usually requires a large volume of data in order to yield accurate result \cite{holzinger2017glass}. However, big data is not always at stake to be used in some problem domains like under-developed languages, clinical trials, biomedical science and etc. Achieving small data machine learning requires optimal utilization of data. Engaging domain experts in the model building process in an interactive way would result in optimal utilization of important data items in a way contributing to the development of accurate models with small data.

To this end, interactive machine learning has a critical role. Hence, it has been studied by various researchers in the area of machine translation \cite{ortiz2010online,ambati2011lowresource_47,gonzalez2012active,peris2017interactive,lam2018reinforcement,foster1997target,huang2021transmart}, computer vision, search engine, social network analysis, music and games, and in data constrained sectors like health.

As it is mentioned above,iML based algorithms defines the state-of-the-art of machine translation systems. In such kind of systems, the knowledge of a human translator is
combined with a MT system. For instance, \cite{ambati2011lowresource_47} propose a resource constrained machine translation. This paper explored active learning as an efficient way to reduce costs and make best use of human resource for building low-resource machine translation systems. Specifically, the author extended the traditional active learning approach of single annotation optimization to handle cases of multiple-type annotations. Furthermore, it show further reduction of costs in building low-resource machine translation systems. Reduction of the required annotated data inherently enabled low resource machine learning. Similarly, various classical IMT researches has been conducted in the few consecutive years. For instance, \cite{ortiz2010online} propose an interactive online machine translation system that avoids the use of batch learning. In \cite{gonzalez2012active}, an AL framework for interactive machine translation specially designed to process data
streams with massive volumes of data is presented. Recently, a couple of researches have also began to incorporate contemporary machine learning techniques for iMT. In \cite{peris2017interactive} and \cite{lam2018reinforcement}, an interactive translation system, based on neural machine translation is deployed. Besides, a full-fledged iMT tools like \cite{huang2021transmart} has been proposed to further boost accuracy of machine translation systems.

Interactive machine learning also plays a significant role in image processing and computer vision in general.The use of iML for automatic feature selection is proposed in \cite{fails2003iml_2}. In this work, the researchers exploit the capacity of iML for automated feature selection. It show how to use iML to eliminate a manual feature selection. The training component of their architecture incorporates the feature selection sub component which further cooperates with the classifier, user feedback manager, and the user as a manual feedback provider. Furthermore, this research introduced an iML image processing framework called Crayons. Crayons is a toolto review and correct classifier errors by using customized decision tree as a learning algorithm and Mean Split Sub Sampled (MSSS) sampling technique.

Search engine optimization is also among the areas that has been studied by iML researchers. In a research contribution by

Besides, \cite{fogarty2008cueflik}, an interactive KNN based desktop application that enable re-rankings on a keyword-based web image search engine relying on the visual characteristics of images is proposed.

Another important iML enabled low resource learning is employed for pattern mining research problem. As it is known, pattern mining is an important process in exploratory data analysis. Various tools have been built to mine patterns from large datasets. However, the problem of identifying patterns that are genuinely interesting to a particular user remains challenging as it requires machine learning experts in addition to the domain expert \cite{dzyuba2014interactive}. To this end, iML plays a vital role both in enabling low resource pattern mining and assuring its valuability for specific users. For instance, \cite{amershi2012regroup_20} formulated a tool called ReGroup that provides users with filters that were generated based on features in the model. Put succinctly, ReGroup leverages the iML paradigm to assist users in creating custom contact groups in online social networks. The classifier provides a potential pattern of users (filters) in social network that likely qualifies for membership in a custom group that is being created. The users are set to provide feedback on each recommended filters where a selection of a contact for inclusion is taken as a positive sample and the remaining are skipped as negative samples. Likewise, the user refines the behaviour of the model with incremental improvements derived through iteratively applied user inputs. Participants noted that these filters provided insight in the patterns that were being exploited by the model, and thus served the dual purpose of explaining the model as well as their intended function as an interaction element. Besides, another research \cite{dzyuba2014interactive} use iML to formulate a pattern mining framework to mine user specific patterns. In this framework, the user is asked to rank small sets of patterns while a ranking function is inferred from this feedback by preference learning techniques. With a user supplemented by active learning heuristics, the resulting accurate rankings of patterns are used to mine new and more interesting patterns. They demonstrated their framework in frequent item-set and subgroup discovery pattern mining tasks. The ability of the framework to learn patterns accurately and the importance of its heuristic approach to ease users effort is presented precisely.

iML has also been applied to various game and entertainment research problems. A sketch-rnn,
\cite{ha2017neural}, present an interactive recurrent neural network (RNN) that help to construct stroke-based drawings of common objects. Besides their model encode existing sketches into a latent vector, and generate similar looking sketches conditioned on the latent space. Similarly, a research work in \cite{jain2020algorithmic} emphasize the design of a deep reinforcement learning agent that can play from feedback alone. This algorithm takes advantage of the structural characteristics of text-based games. Moreover, the application of iML in motion-driven music systems is presented in \cite{fiebrink2010wekinator_12,gillian2014gesture,schedel2011wekinating}.Due to the complications to design a robust player-recognition or motion recognition system using standalone iML system, an enhanced model has been presented in \cite{diaz2019interactive}. They proposed an iML solution for Unity3D game engine in the form of a visual node system supporting classification (with k-nearest neighbour), regression (with multilayer perceptron neural networks) and time series analysis (with dynamic time warping) of sensor data.

The health sector is also known for data scarcity. In this sector, researchers are often confronted with only a small number of dataset or rare events, where a machine learning algorithm suffers from insufficient training samples \cite{holzinger2016interactive, maadi2021review}. To this end iML researches has been conducted to build resource constrained model that adheres to the trustworthiness principles. In \cite{Holzinger2018InteractiveML_9}, an interactive Ant Colony Optimization (ACO) technique is applied to the Traveling Salesman Problem (TSP). The TSP is an intransigent mathematical problem to find the shortest path through a set of points and returning to the origin. It is among the problems that resembles most computational problems in the health informatics sector. They introduce two novel concepts: Human-Interaction-Matrix (HIM) and Human-Impact-Factor (HIF) to control the ants action and the variable interpretation of the HIM respectively. The authors implemented their framework using a java-script based browser solution which has a great benefit of platform independence and configuration. Although the results they found are promising, the researchers suggested further works like such as gamification and crowdsourcing to solve hard computational problems.

As we have discussed in the first section of this document, the essence of iML is to leverage the benefits of the human-in-the-loop. However, the presence of the human does not by itself guarantee the trustworthiness and performance of the model. The interaction between the user and the learning model should be supplemented with clear and in-depth visualization methods. To this end, \cite{amershi2011effective} emphasize on how to design effective end-user interaction with interactive machine learning systems. They shape the human-model interaction in a way it answers these critical questions: which examples should a person provide to efficiently train the system, how should the system illustrate its current understanding and how can a person evaluate the quality of the system’s current understanding in order to better guide it towards the desired behavior. Another practical contribution for human-model interaction is provided by \cite{li2018interactive}. They propose a new visual analytic approach to iML and visual data mining. Multi-dimensional data visualization techniques are employed to facilitate user interactions with the machine learning and mining process. This allows dynamic user feedback forms (data selection, data labeling and correction) to enhance the efficiency of model building. In particular, this approach can significantly reduce the amount of data required for training accurate models. Hence, it can be highly impactful for applications where large amount of data is hard to obtain. The proposed approach is tested on two application problems: the handwriting recognition (classification) problem and the human cognitive score prediction (regression) problem. Both experiments show that visualization supported iML and data mining can achieve the same accuracy as an automatic process can with much smaller training data sets. Furthermore, \cite{arendt2018interactive} propose a scalable client-server system for real-time iML framework called Computer-Human Interaction for Semi-Supervised Learning (CHISSL) \cite{arendt2017chissl}. The proposed system is capable of incorporating user feedback incrementally and immediately without a predefined prediction task. The light-weight computation web-client and heavyweight server constitute their architecture. The server relies on representation learning and off-the-shelf agglomerative clustering to find a dendrogram, which we use to quickly approximate distances in the representation space. The client, using only this dendrogram, incorporates user feedback via transduction. This work achieves low resource learning as it updates distances and predictions for each unlabeled instance incrementally and deterministically, with O(n) space and time complexity. Ultimately, this paper solved the scalability issue of CHISSL. Another important contribution in this regard is that of \cite{talbot2009ensemblematrix}. In this paper, EnsembleMatrix, SVM based visualization system, to tune classifier through user in the human-in-the-loop approach to build a superior model. This product presents a graphical view of confusion matrices to help users to directly interact with the visualization in order to explore and build combination models. 

\subsection{Pervasive Machine Learning}
Interactive machine learning can also yield a better performance in pervasive computing environment where there are low computing resources. For instance, \cite{frazier2019improving} present techniques to enable low resource computation for deep reinforcement learning agents complex behaviors in 3D virtual environments. Specifically, they considered an environment with high degree of aliasing, MineCraft, and conducted experiments with two reinforcement learning algorithms which enable human teachers to give advice-Feedback Arbitration, and Newtonian Action Advice under visual aliasing conditions. Similarly, \cite{preuveneers2020resource_ref17}  emphasizes on the importance of hyper tuning as a process to select the best performing machine learning model, its architecture and parameters for a given task. The fact that hyper parameter tuning does not take into consideration resource trade-offs when selecting the best model for deployment in smart environments has been taken as their research challenge. Consequently, the authors proposed a multi-objective optimization solution to find acceptable trade-offs between model accuracy and resource consumption to enable the deployment of machine learning models in resource constrained pervasive environments. On the other hand, \cite{malloch2017fieldward} used pathward and fieldward techniques to enable creation of gestures that can be used by a touch-enabled pervasive mobile devices reliably. They introduced a novel dynamic guide that visualizes the negative space of possible gestures as the user interacts with the system.

On the other hand, \cite{tegen2020activity_16} emphasises on the gaps of standard machine learning algorithms in dynamic sensor setting. Consequently, an iML based framework is  implemented for “activity recognition” problems. This framework takes stream of data among different devices in the internet of things as an input. Three different machine learning approaches were used in the experiments: Support Vector Machine, k-Nearest Neighbor and Naıve Bayes classifier. These algorithms were customized to adapt the dynamic sensory setting. Uncertainty, Error (output from the model), State Change (what change in state urges the user to change label of data), Time (the time difference b/n queries and labeling by user), and Randomness (randomly assigning labels) are identified as factors that increase the subjectivity of users feedback. The results of this research work make it clear that the choice of interactive learning strategy has a significant effect on the performance when tested on recordings of streaming data.

\section{iML Performance Evaluation}
\label{imleval}
Although there are a number of invaluable contributions towards the goal of iML algorithms, the techniques used to evaluate their performance are not that mature. The standard machine learning algorithms use statistical model performance evaluation techniques which provide only a sole measure, obfuscating details about critical instances, failures and model features \cite{ma2019iml_46}. However, the "human-in-the-loop" nature of iML systems triggers an extended need to diagnose the subjectivity of results. Therefore, iML performance evaluation should consider subjectively generated user evaluations and the cyclic nature of influence between the algorithm and users. This makes the evaluation of iML systems subjective and complex \cite{boukhelifa2018evaluation_3, yang2019evaluating_45}.

Consequently, researchers proposed both customized and noble solutions to approach these critical issues. The iML Framework for Guided Visual Exploration (EvoGraphDice) \cite{boukhelifa2018evaluation_3} is among the salient ones that is produced after an extensive experimental review of existing evaluation techniques. This paper founded its base mainly on the gaps of user based and system based feedback evaluation techniques to get insights on the underlying co-operation and co-adaptation mechanisms between the algorithm and the human. EvoGraphDice couples algorithm-centered and user-centered evaluations to bring forth insights on the underlying co-operation and co-adaptation mechanisms between the algorithm and the Human.
Other papers focus on the input and output nature of iML models to diagnose their performance. For instance, the INFUSE system \cite{krause2014infuse_ref19} supports the interactive ranking of features based on feature selection algorithms and cross-validation performances. Another work \cite{krause2017workflow_ref18} also propose a performance diagnosis workflow. In this work, the instance-level diagnosis leverages measures of “local feature relevance” to guide the visual inspection of root causes that trigger misclassification.

\cite{yang2019evaluating_45} identify diversified scenarios and subjective nature of explanations for the absence of benchmarks to evaluate explanations of iML algorithms.Consequently, they have defined the problem of evaluating explanations and systematically reviewed the existing efforts from state-of-the-arts. The authors discuss explanation as global (overall working structure) and local (particular model behaviour for individual instances) from scope perspective and as intrinsic (self-interpretable models) and posthoc (independent interpretation model) from dimension perspective. Furthermore, generalizability (generalization), fidelity (degree of exactness) and persuasibility (comprehensibility) are identified as the three properties of explanation and corresponding methods has been revised for each aspect. As a result, they designed a unified evaluation framework based on the discussed properties of explanation and to the hierarchical needs from developers and end-users. In this work, three tiered explanation evaluation architecture is proposed corresponding to the three properties of explanation.

A similar work \cite{felderer2021quality} discusses the importance of enhancing the quality of AI-based systems for a practical usage. They affirmed the importance of quality assurance to set benchmark for evaluations. Understandability and interpretability, defining expected outcomes as test oracles, and non-functional properties of AI-based systems of AI models are among the seven challenges they listed to assure quality of AI-based systems.

\section{Discussion}
\label{Potential}

An extensive merit-oriented state-of-the-art review of iML is presented in this paper. We used a bottom-up approach to categorize researches based on their primary importance. As shown in Table 1, robust machine learning, trustworthy machine learning, and low resource machine learning are identified as the major themes to categorize research works contributing to the state-of-the-art of iML. Significant issues related to iML evaluation has also urged us to consider it as one of the defining factors of the state-of-the-art of iML. 

Furthermore, these research works have been further classified under their technical and sectoral emphasis. We subsume research works with specific component-wise and task-wise concentration under the technical section. Specifically, as shown in Table 1, HCI and Interface design, iML visualization, and model explanability related research works constitute the partial taxonomy of the technical section. Moreover, information processing class enclose research works focusing on the use of iML for training data management in areas such as machine translation. Searching and retrieval section contains iML inspired research works for search engine and retrieval systems. Adversarial attacks and iML inspired defense strategies are put under security and privacy section. Interactive machine learning inspired researches for low resource (pervasive) environment and clustering and optimization problems are classified under the technical section of the merit-oriented taxonomy accordingly. On the other hand, papers with special emphasis on general sectors like agriculture, health, game and development, and education are put under the sectoral section of the taxonomy. Finally, research works that focus on multiple classes and are generic to the topic we are studying are put under the other section of our iML taxonomy. 

In the following sections, we discuss the major pillars of this state-of-the-art review. Moreover, the gaps identified in the revised research works are discussed as research opportunities. 

\begin{figure}[htbp]
  \centering
  \includegraphics[width=\linewidth]{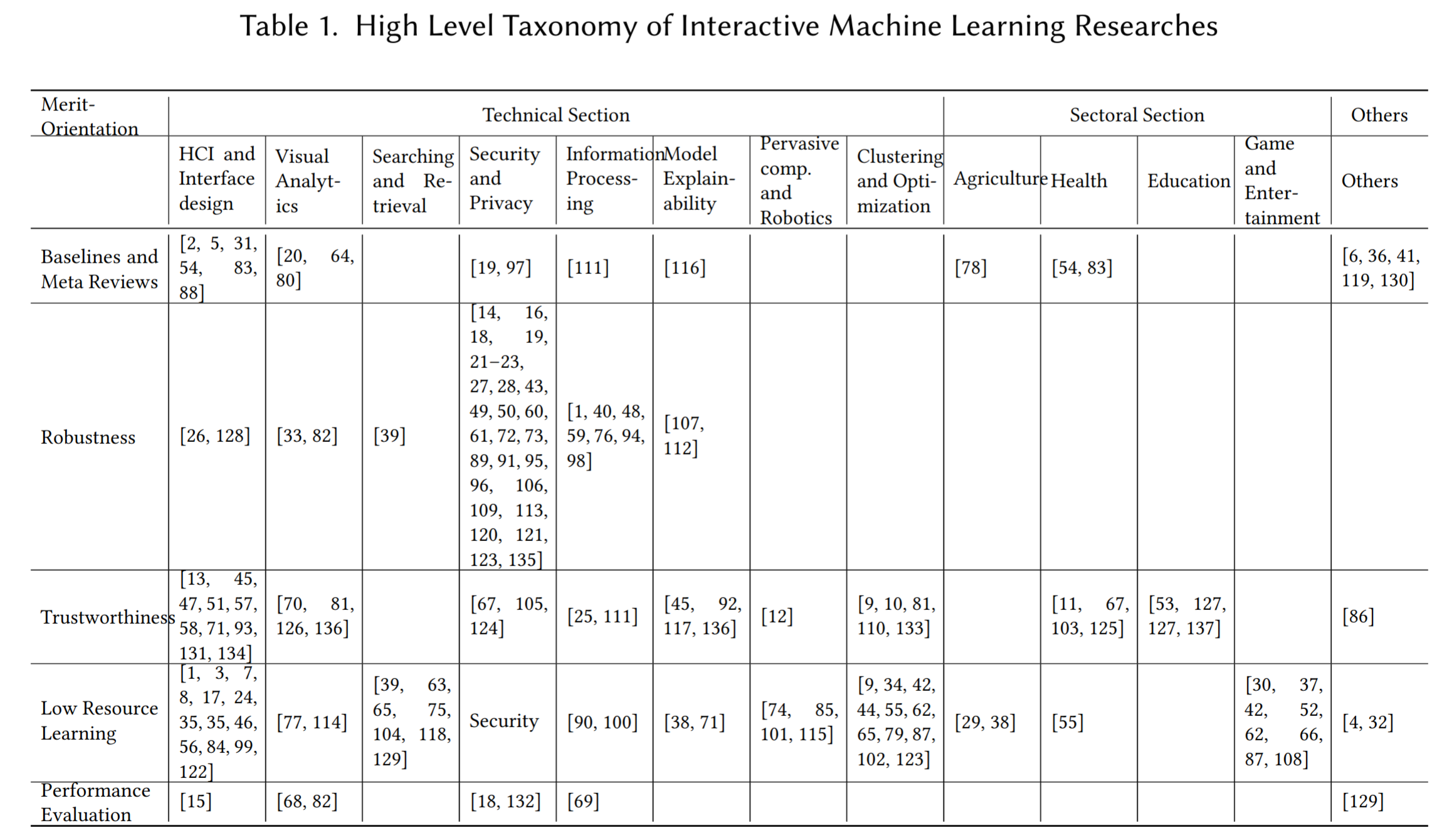}
  \caption{Merit-Oriented Architecture of Interactive Machine Learning}
  \label{imlsummary}
\end{figure}

\subsection{ Robust Machine Learning}
Tackling adversarial attacks has been one of the salient research areas in the areas of machine learning. Consequently, a number of adversarial example attacks have been discovered with the intention of exposing potential gaps on the learning models. However, such kind of approaches are reactive to solving robust machine learning.

From the attack surface perspective \cite{papernot2016towards}, the nature of input perturbation \cite{chen2017targeted, zhang2019poisoning, geiping2021doesn, carlini2017blackbox_ref9,szegedy2013blackbox_ref10}, test perturbation \cite{biggio2013evasion, kwon2018friend, kwon2018multi}, and model exploitation \cite{pappernot2017adv_ref4, brendel2017blackbox_ref8, guo2019blackbox_ref11} has been studied extensively.

As it is thoroughly discussed from section \ref{robust}, iML has been extensively applied to address the problem of model robustness. The methods against adversarial example attacks mainly basis on visualization of either the models intermediate output, data instances, local structures \cite{ma2019iml_46, emamjomeh2017general} or the adversarial attacks \cite{das2020massif, morris2020textattack}, their masking techniques \cite{slack2020fooling} and visualization of the system's explanation \cite{das2018adagio}.

Incorporation of an interactive component in the model building process is experimentally proved to have positive impact on tackling input perturbation. Human-in-the-loop can help to identify intentionally created parametric and data instance uncertainties. Similarly, users are observed to have successfully detecting either manipulated neurons (suppressed or hyper-activated) or data instances. Accordingly, the human-in-the-loop feature of iML evidently contributes in detecting adversarial attacks leading to the early enhancement of model robustness possible.

Given the increased complexity of adversarial attacks over-time and its subjective nature, the iML inspired research outcomes at hand falls short to sustainably address the problem. The power of model visualization and users knowledge towards advanced adversarial attacks should be studied to better solve the problem at hand. 

\subsection{Trustworthy Machine Learning}

The most critical setback machine learning industry faces in recent years is a hampered applicability for decision sensitive sectors like Health, Finance, Military, and Autonomous Robotics like autonomous driving. Scientists hesitate to use standard machine learning for many trust-related reasons. The black-box nature of models, less control of users over data instances and absence of guidelines to formalize the integration of exploratory components in standard machine learning are some of the reasons behind its hampered application. Besides, the relatively bigger error margin of standard machine learning algorithms and the small error sensitivity of models in the aforementioned sectors makes it hard to confidently employ standard machine learning algorithms.

To this end, we have reviewed research conducted to enhance trustworthiness of models. Put succinctly, we have reiterated on the contribution of iML algorithms in enhancing the trustworthiness of iML algorithms.

Most iML inspired techniques utilizes state-of-the-art textual  \cite{Mozina2018explanatory_7, turchetta2019safe} and visualization components \cite{van2011iml_ref14,liu2017iml_ref15,zhao2018iml_ref16, Muhlbacher2014blackbox_ref13,van2011iml_ref14,ambati2011lowresource_47, Stepano2019Explanatory_6,liu2017iml_ref15, zhao2018iml_ref16}.

The salient works in the textual explanatory aspects of iML reiterates mainly on deriving techniques to explain and interpret intermediate results of a model textually. The effectiveness of explanations mainly depends on their degree of interpretability by the users. Techniques like Argument based Axiomatization \cite{Mozina2018explanatory_7} and goal-oriented safe-exploration \cite{turchetta2019safe} has been reviewed as an important mechanisms to enhance the explanatory capability of models in general. On the other hand, visual exploratory methods are also found to provide better insights to complex models. Visualizations have been performed in either of intermediate results \cite{Stepano2019Explanatory_6} or the explanations themselves \cite{liu2017iml_ref15,zhao2018iml_ref16}. Correlations between data, intermediate results and patterns among intermediate results have been used for visualization. Visualizations has also been applied to different data structures. We have seen techniques to visualize decision tree structures \cite{van2011iml_ref14} and tree ensemble models through the use of multiple coordinated views \cite{liu2017iml_ref15,zhao2018iml_ref16}.

From the application point of view, it is demonstrated that iML inspired neuro-evolutionary algorithms boost trusts among autonomous agents \cite{robert2017Robot_5}. Similarly, iML can also be used to increase the trustability of models in the health sector \cite{Holzinger2018InteractiveML_9}. For instance, \cite{kulesza2015principles} developed an explanation-centric approach to help end users to effectively and efficiently personalize machine learning systems. It intends to increase end users trust to a learning model by exposing the details of a learning model using an Explanatory debugging technique.

The other significant problem we have examined from the model trustworthiness perspective is absence of guidelines to build explanatory components. The employment of off-the-shelf interface design techniques resulted in poor performance of the overall algorithm \cite{bernardo2017interactive}. To this end, some works \cite{Muhlbacher2014blackbox_ref13} are conducted to set guidelines for integrating visualization into machine learning algorithms.

Besides, improving users experience can also boost its trustworthiness. In \cite{bernardo2017interactive}, the importance of iML to improve user experiences by supporting user-centred design processes is studied. Besides, they conclude that there is a further role for user-centred design in improving interactive and classical machine learning systems.

From what we have revised, iML plays a vital role in introducing interactive learning and explainability.The fundamental assumption in this regard is that the more transparent model gains higher degree of trust. However, some researches have also tried to hypothesize an inverse relationship between trust and explainability \cite{honeycutt2020soliciting}. Incidences like this might happen if users don't possess better understanding of black-box models and as a result focus on the uncovered error margins of transparent models. Similar findings reaffirm the importance of undergoing empirical model evaluations in addition to subjective ones in order to decide the trustworthiness of a given model. 

Therefore, further researches on exploratory machine learning should be among the salient research problems in the iML domain. Besides, the design of user experience for iML should be studied further as it contributes for the effectiveness of the human-in-the-loop modality. 

\subsection{Low Resource Machine Learning}
Machine learning consumes large data and resources in order to produce accurate models and undergone efficient performance evaluation. Generally, it is common knowledge that too little training data results in poor approximation. For instance, no human would need several thousand images to learn numbers in another language, yet a fully connected neural network needs too much data for good classification of MNIST dataset. On the contrary, too little test data will result in an optimistic and high variance estimation of model performance. 

The amount of training data for machine learning has direct relationship with computing resources used to build models. The more data poured to the model builder, the more computational resource it requires. As a result, application of standard machine learning in resource constrained environments, more specifically in pervasive and robotics environments, has become less realistic. Sectors like under-developed linguistics, medical science, finance and military may not have sufficient amount of data for the standard machine learning framework to produce accurate outcome. Reducing the amount of data in these type of sectors compromises quality as the domain is sensitive to even smaller error margins.

Consequently, machine learning algorithms has been modified in a way it engages users-in-the-loop. Besides, it incorporates state-of-the-art visualization techniques to leverage users feed backs for data reduction and easing computing resource requirements. To this end, iML plays a vital role. 

As we have clearly analysed in the state-of-the-art review, iML takes advantage of users feedbacks on intermediate results to reduce the amount of data and computational resources required to build well performing model. To this end, various iML inspired researches have been conducted. iML tools has been employed to enable low resource feature selection \cite{fails2003iml_2}, Activity recognition in dynamic sensor setting \cite{tegen2020activity_16}, pattern mining and matching in social network analysis \cite{dzyuba2014interactive}, data exploitation \cite{amershi2012regroup_20} and other applications. iML, especially its explanatory components, has been used to take advantage of some computing architectures. For instance, as we have precisely covered in this document, integration of iML to  Deep Reinforcement Learning (DRL) has been experimented for 3D virtual environments \cite{frazier2019improving}. This shows how well researchers recognized the importance of iML for resource constrained learning. 

However, the contributions does not yet suffice the problems at hand. That means, iML is not exploited enough for critical demands of low resource learning especially in the area of robotics and pervasive computing. Robot simulation \cite{marco2017virtual}, low resource object recognition and tracking, navigation and natural language capabilities of robots can be enhanced using iML inspired techniques. Therefore, exploiting iML for low resource machine learning especially in the area of machine intelligence and simulation is identified as a promising area of research. 

\subsection{iML Performance Evaluation}
iML comes along with its complications of performance measure. Performance of iML inspired algorithms can not be effectively measured for two critical reasons namely: the subjectivity of users feedback and the co-operation and co-adaptation or the two sided influence between users and the system. The subjective nature of users in iML inspired model building makes it hard to put benchmarks for model evaluation \cite{yang2019evaluating_45}. On the other hand, system based feedbacks are prone to ill interpretability. Consequently, the empirical performance measuring techniques of standard machine learning can not be exclusively used to assure the quality of iML inspired algorithms. 

To this end some researches tried to address performance evaluation problems by addressing user's and systems feedback separately. Both the subjectivity of users feedback and degree of interpretability of systems explanation for a given intermediate results contributes to the complex nature of iML evaluation. Therefore, a partial effort to address one of these problems won't provide the desired solution. Similarly, focusing on user-centeric performance evaluation techniques does not suffice the problem at hand. 

To this end, a coupled performance evaluation technique that utilizes both subjective and empirical performance evaluation techniques happens to yield better results \cite{boukhelifa2018evaluation_3}. Interactive ranking of features \cite{krause2014infuse_ref19} and more specifically diagnosis of instances to leverage measures of local feature relevance are studied by different researchers. However, this area is still open for further research as there is no generalized and robust performance evaluation technique for iML as there is for statistical machine learning algorithms.

\section{Conclusion and Future Works}
We have seen how iML presents interesting challenges and prospects to conduct future research; not only in terms of designing robust algorithms and interaction techniques, but also in terms of coherent evaluation methodologies. Consequently, an extensive review of the state-of-the-art of iML is conducted in this paper.

From the tabular representation of iML inspired research works, shown in Table 1, it is clear that iML has not been exhaustively applied to areas such as searching and retrieval, pervasive computing and robotics, and clustering and optimization tasks. Moreover, although the importance of iML in decision-sensitive areas such as agriculture, health, education, game and entertainment is evidential, we have seen very few contributions in this regard.

Consequently, we have come to a conclusion that the domain of machine learning has not yet exploited the power of iML features and the human exhaustively. The merits of iML have not yet been fully enriched for contemporary machine learning algorithms. Identifying gaps and approaching them with state-of-the-art methods should be the primary goal iML researchers. To this end, we summarize potential research problems that we believe are inspiring for future work in iML as follows: 
\subsection{Adversarial Attack Defense}

As per our analysis under the section of robust machine learning, the primary source of adversarial attacks is black-box nature of algorithms and bulk use of data. Interactive machine learning should further be studied in this aspect as it is evidently an efficient approach for small data and exploratory machine learning.

Further researches should be conducted to leverage the human-in-the-loop feature and suppress any training-phase and test-phase adversarial attacks. The engagement of the user in the model building pipeline will presumably reduce the degree of severity adversarial attacks against machine learning models.

\subsection{Degree of User Engagement}

The degree of users engagement in an interactive model building process is not well-defined. This may detrimentally affect model accuracy and overall performance as clearly presented in \cite{honeycutt2020soliciting}. Therefore, a separate effort to set the degree of user engagement while building robust and trustworthy models should be researched thoroughly.

\subsection{Exploratory Machine Learning}

As we have discussed in the previous sections, empowering machine learning algorithms with exploratory features is a fundamental step towards transparent, trustworthy, and accurate model based computation.

However, providing explanations and interpretations on intermediate results of a model a critical problem until now. Not much work has been done in integrating iML features with contemporary machine learning algorithms. Therefore, we recommend further researches in building exploratory machine learning frameworks.

\subsection{Low Resource Learning}

Resource constrained learning is among the critical problems a machine learning industry face these days. The increased pervasiveness of computing nodes and the scarcity of data affected the applicability of machine learning significantly. Besides, the need for real time learning increased overtime. One important way to overcome this problem and ensure the applicability of machine learning in low resource machines is a systematic employment of iML. Especially, as we it is seen in our analysis, more effort on iML would enable real time learning on robots and other low resource agents\cite{mordatch2015interactive}. Furthermore, complex tasks like object detection, recognition and tracking, saliency identification, face and emotion recognition, natural language processing, robot navigation and simulation can be significantly enhanced with less data and computational resource by using interactive machine learning.

\subsection{User Experience}

The essence of interactive machine learning to engage users in the model building process and as a result obtain advantage of it. This can only be achieved if we provide a good user experience for the user subject to advise the learning model. To this end, user experience design plays a vital role.

Designing user interfaces using state-of-the-art design methods and in conformance with the specific requirements of iML algorithms \cite{dudley2018review_4}. This allows experts outside the machine learning domain use machine learning algorithms to solve problems at their domain of expertise without much agony.

\subsection{iML Performance Evaluation} Performance evaluation is one of the complications brought to the world of machine learning along with iML. The subjective, co-operation and co-adaptation nature of iML makes it hard to set benchmarks for performance evaluation. Therefore, we suggest further researching to enhance performance evaluation techniques for iML.

\section{Acknowledgements}

This work would not have been possible without the financial support of the French Embassy in Ethiopia, Brittany region administration and the Ethiopia Ministry of Education (MoE). We are also indebted to Brest National School of Engineering (ENIB) and specifically LABSTICC for creating such a conducive research environment.

\bibliographystyle{unsrtnat}
\bibliography{references}  






\end{document}